\documentclass[conference]{IEEEtran}
\IEEEoverridecommandlockouts
\usepackage{cite}
\usepackage{amsmath,amssymb,amsfonts}
\usepackage{algorithmic}
\usepackage{graphicx}
\usepackage{textcomp}
\usepackage{soul}
\usepackage{multirow}

\usepackage{xcolor}
\usepackage{colortbl}

\definecolor{Gray}{gray}{0.9}
\definecolor{DarkGray}{gray}{0.8}
\definecolor{LightGray}{gray}{0.97}

\def\BibTeX{{\rm B\kern-.05em{\sc i\kern-.025em b}\kern-.08em
    T\kern-.1667em\lower.7ex\hbox{E}\kern-.125emX}}
\begin{document}

\title{Real-World Community-in-the-Loop Smart Video Surveillance - A Case Study at a Community College\\
\thanks{This research is supported by the National Science Foundation (NSF).}
}

% \author{Anonymous Submission}

\author
{\IEEEauthorblockN{1\textsuperscript{st} Shanle Yao}
\IEEEauthorblockA{ \textit{Electrical and Computer Engineering Department}\\
 \textit{The University of North Carolina at Charlotte}\\
 Charlotte, USA \\
 syao@uncc.edu}
\and
\IEEEauthorblockN{2\textsuperscript{nd} Babak Rahimi~Ardabili}
\IEEEauthorblockA{\textit{Public Policy} \\
\textit{The University of North Carolina at Charlotte}\\
Charlotte, USA \\
brahimia@uncc.edu}
\and
\IEEEauthorblockN{3\textsuperscript{rd} Armin Danesh~Pazho}
\IEEEauthorblockA{\textit{Electrical and Computer Engineering Department}\\
\textit{The University of North Carolina at Charlotte}\\
Charlotte, USA \\
adaneshp@uncc.edu}
\and
\IEEEauthorblockN{4\textsuperscript{th} Ghazal Alinezhad~Noghre}
\IEEEauthorblockA{\textit{Electrical and Computer Engineering Department}\\
\textit{The University of North Carolina at Charlotte}\\
Charlotte, USA \\
galinezh@uncc.edu}
\and
\IEEEauthorblockN{5\textsuperscript{th} Christopher Neff}
\IEEEauthorblockA{\textit{Electrical and Computer Engineering Department}\\
\textit{The University of North Carolina at Charlotte}\\
Charlotte, USA \\
cneff1@uncc.edu}
\and
\IEEEauthorblockN{6\textsuperscript{th} Hamed Tabkhi}
\IEEEauthorblockA{\textit{Electrical and Computer Engineering Department}\\
\textit{The University of North Carolina at Charlotte}\\
 Charlotte, USA \\
 htabkhiv@uncc.edu}
}

\maketitle

\begin{abstract}
Smart Video surveillance systems have become important recently for ensuring public safety and security, especially in smart cities. However, applying real-time artificial intelligence technologies combined with low-latency notification and alarming has made deploying these systems quite challenging. 

This paper presents a case study for designing and deploying smart video surveillance systems based on a real-world testbed at a community college. We primarily focus on a smart camera-based system that can identify suspicious/abnormal activities and alert the stakeholders and residents immediately. The paper highlights and addresses different algorithmic and system design challenges to guarantee real-time high-accuracy video analytics processing in the testbed. It also presents an example of cloud system infrastructure and a mobile application for real-time notification to keep students, faculty/staff, and responsible security personnel in the loop. At the same time, it covers the design decision to maintain communities' privacy and ethical requirements as well as hardware configuration and setups. 

We evaluate the system's performance using throughput and end-to-end latency. The experiment results show that, on average, our system's end-to-end latency to notify the end users in case of detecting suspicious objects is 5.3, 5.78, and 11.11 seconds when running 1, 4, and 8 cameras, respectively.
On the other hand, in case of detecting anomalous behaviors, the system could notify the end users with 7.3, 7.63, and 20.78 seconds average latency. These results demonstrate that the system effectively detects and notifies abnormal behaviors and suspicious objects to the end users within a reasonable period. The system can run eight cameras simultaneously at a 32.41 Frame Per Second (FPS) rate. 

\end{abstract}

\begin{IEEEkeywords}
smart video surveillance, case study, anomaly detection, public safety 
\end{IEEEkeywords}

\section{Introduction}

Video surveillance systems have long been recognized as a tool for monitoring and ensuring public safety and security\cite{17}. However, recently with the rapid advancements in machine learning algorithms, both in hardware and software, smart video surveillance (SVS) systems are becoming more pervasive in public, and private sectors \cite{18}. According to market research reports, the global video surveillance industry achieved a valuation of USD 19.12 billion in the year 2018. This market is anticipated to expand at a compound annual growth rate (CAGR) of 6.8 during the forecast period, ultimately reaching a projected valuation of USD 33.60 billion by the year 2026\footnote{https://www.fortunebusinessinsights.com/video-surveillance-market-102673}. The increasing demand for smart video surveillance cannot be solely attributed to market interests, as the escalating safety concerns in contemporary society also propel it. The need for enhanced security measures has become a pressing priority for individuals and organizations alike, thereby driving the adoption of this technology. According to Gun Violence Archive, 108 mass shootings in the USA have been recorded as of March 11, 2023 \footnote{https://www.gunviolencearchive.org/}. Mass shootings have not been limited to regions with high crime rates; several high-profile mass shootings in supposedly safe public locations such as schools and universities have been reported in recent years. Michigan State University, Robb Elementary School in Uvalde, Texas, and Westinghouse Academy in Pittsburgh, Pennsylvania, are the most current and disastrous incidents that show SVS adaptation's importance in public places.

\begin{figure}
    \centering
       \includegraphics[width=1\linewidth, trim= 0 150 0 50,clip]{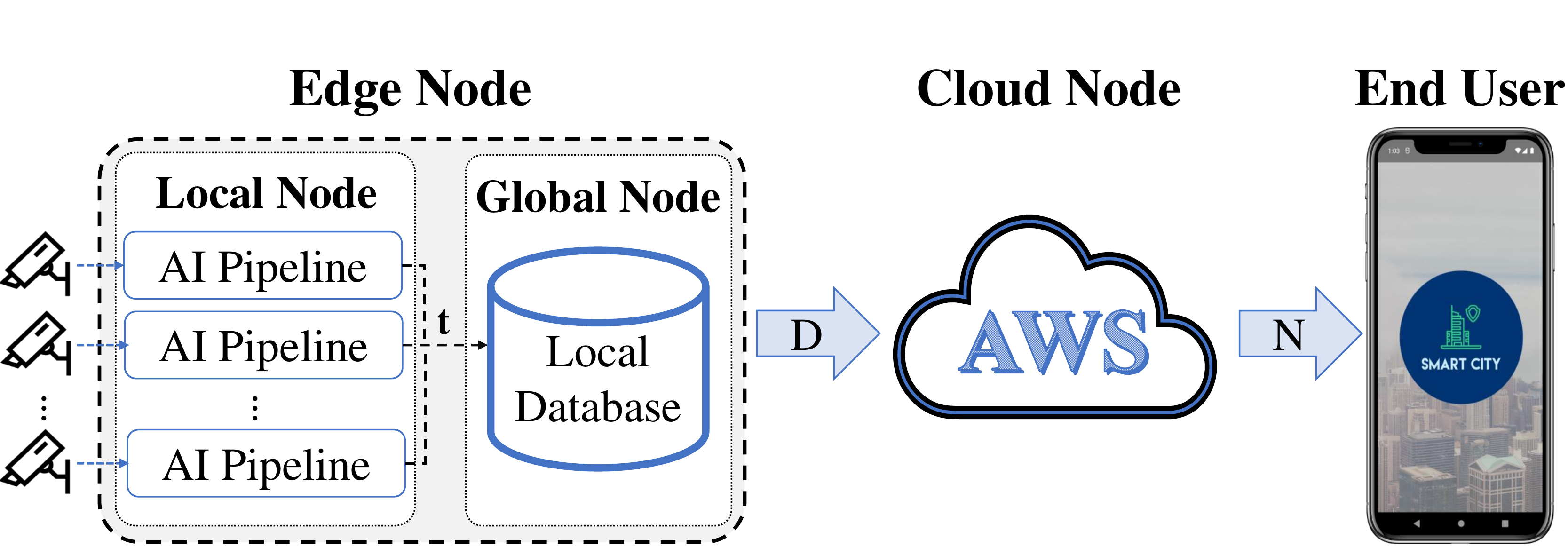}
            \caption{Conceptual Model.}
            
        \label{fig:Concept}
\end{figure} 

Deploying smart video surveillance systems has become a viable solution for enhancing the capabilities of traditional surveillance systems while assisting law enforcement \cite{19} and \cite{20}. In the design of a smart video surveillance (SVS) system, particular attention should be given to its ability to detect anomalies promptly and notify the end-users in the event of suspicious incidents. This is critical to ensure the system's effectiveness in enhancing security and mitigating potential threats \cite{21}. The scalability issue also poses a significant challenge in deploying an SVS system. \cite{22}.  

As discussed, using SVS systems has become more prevalent in recent years to enhance security in public places. However, there are concerns about the potential privacy threats of such systems \cite{23}, mainly if they are implemented without community involvement or consultation. As such, the "community in the loop" concept has emerged as a critical consideration in deploying AI-driven technologies \cite{24} such as SVS systems. Engaging with the local community and involving them in the planning and implementation of SVS systems is essential in establishing trust and acceptance of such systems\cite{25}. Doing so can help ensure that the system is used ethically and that the privacy of individuals is protected. Additionally, community involvement can help identify the most pressing security concerns and ensure the system is designed and used to address these concerns effectively\cite{25}. Therefore, ensuring that the system is connected to the community to receive the outputs of an SVS system is crucial in designing such a system. 

To mitigate the challenges mentioned earlier in a practical setting, a case study was conducted at a Community College wherein an SVS system was implemented to monitor and identify anomalies in the surveillance footage. The selection of the community college as the testbed was intentional. It aimed to address the challenges above in an environment with a comparatively lower likelihood of anomalous behavior than in high crime rate environments.
In this study, we utilized a designed scalable SVS system \cite{2} to conduct the Artificial Intelligence (AI) vision analysis and detect anomalous behaviors. We also employed a smartphone application \cite{15} to push notifications to the end user in case of detecting anomalies. Figure \ref{fig:Concept} represents the conceptual end-to-end model used in this study. 

The study evaluates the end-to-end latency and throughput performance of the SVS system, which can detect object anomalies and behavioral anomalies that commonly occur in a campus setting. The study is conducted by monitoring 1/4/8 cameras simultaneously, utilizing a deep learning pipeline consisting of multiple stages, including image extraction, object detection, person tracking, and behavioral anomaly detection. The performance evaluation of the SVS system is conducted under realistic conditions, using a range of simulated anomalies, to assess the system's ability to detect and respond to potential security threats. 

We assessed the system's performance by measuring its throughput and end-to-end latency. Our findings show that our system can process eight cameras simultaneously at 32.41 Frames Per Second (FPS). The results indicate that the system takes 5.3, 5.78, and 11.11 seconds to detect suspicious objects for 1, 4, and 8 camera setups. For detecting anomalous behavior, the system notifies end-users with an average latency of 7.3, 7.63, and 20.78 seconds, respectively.

Our main contributions are:

\begin{itemize}
  \item Conducting a case study with a real-world testbed at a community college to address the challenges of practical deployment of an SVS system in a comparatively low-crime environment; 
  \item Evaluating the end-to-end latency and throughput performance of the system as it interacts with the real-world environment; This includes measuring the system's performance in conducting behavioral anomaly detection as a downstream task in real-world;
  \item Providing technical guidelines, which also addresses  ethical and privacy concerns, of real-world deployment of community-in-the-loop SVS.
\end{itemize}

This paper is organized into several sections to provide a comprehensive understanding of deploying a real-world community-in-the-loop SVS case study. Section 2 begins by reviewing the relevant works on this topic. Section 3 describes the system features. Section 4 presents the test setup and results of our experiments. Building on these findings, in section 5, we summarize our findings and outline directions for future research.

\section{Related Works}

The idea and goal of SVS systems are able to notify the crime in real-time and able to trace back the suspicious activity with historical tracking data in recent years study. The Focus of this study is usually separated into two parts: tracking with re-identification and anomaly scene detections for both traffic and humans. There are many works \cite{3} \cite{4} \cite{5} trying to achieve real-time object tracking on edge devices with lightweight models and showing awe-inspiring results which can be implemented in the existing system with low latency easily. Study such as \cite{8} talks about how SVS can be used for traffic safety prevention and addressing abnormal traffic events appearing on the street, including speed violation, one-way traffic, overtaking, illegal parking, and dangerous dropping off of passengers. Different decision trees are assigned in the system for each abnormal event to calculate the classification scores for alerting purposes.

Researchers from the University of Technology Gorakhpur introduced an SVS system \cite{6} in 2023, employing advanced pre-processing techniques on edge devices to improve the accuracy in detecting falling behavior from surveillance cameras. In the preceding year, the E2E-VSDL \cite{13} method was proposed, utilizing BiGRU and CNN to compare the video frames with identified testing anomaly videos in their trained model to classify them as either abnormal or not. The related work achieved approximately 98 percent accuracy with the existing dataset. A novel deep contrastive learning for SVS methodology \cite{10} is published, called TAC-Net using a self-supervised learning model to capture high-level semantic features and tackle anomaly scenes. TAC-Net\cite{10} results in outperforming existing State-Of-The-Art (SOTA) methods on poplar anomaly datasets in 2021.

There are also many other SVS studies trying to move the system into real-world  applications.  Paper\cite{39} published in 2019 actually shows a running system being used with surveillance cameras rather than testing on the dataset. To enhance the performance of anomaly detection models and achieve superior training outcomes, researchers from RV College of Engineering Bengaluru \cite{41} try to improve localization results from object detector models, and it turns out to increase the validation accuracy results to about 98.5 percent.

Optimizing large-scale SVS systems has gained significant importance in recent research, and there are also many great systems \cite{9}\cite{7}\cite{12} focusing on optimizing data transforming, communication, and even incorporating blockchain technology in various domains. These systems not only emphasize machine learning model performance but also consider the larger scalability of the system as a whole. 
 \begin{figure*}[]
        \centering
               \includegraphics[width=1\linewidth, trim= 0 280 5 25,clip]{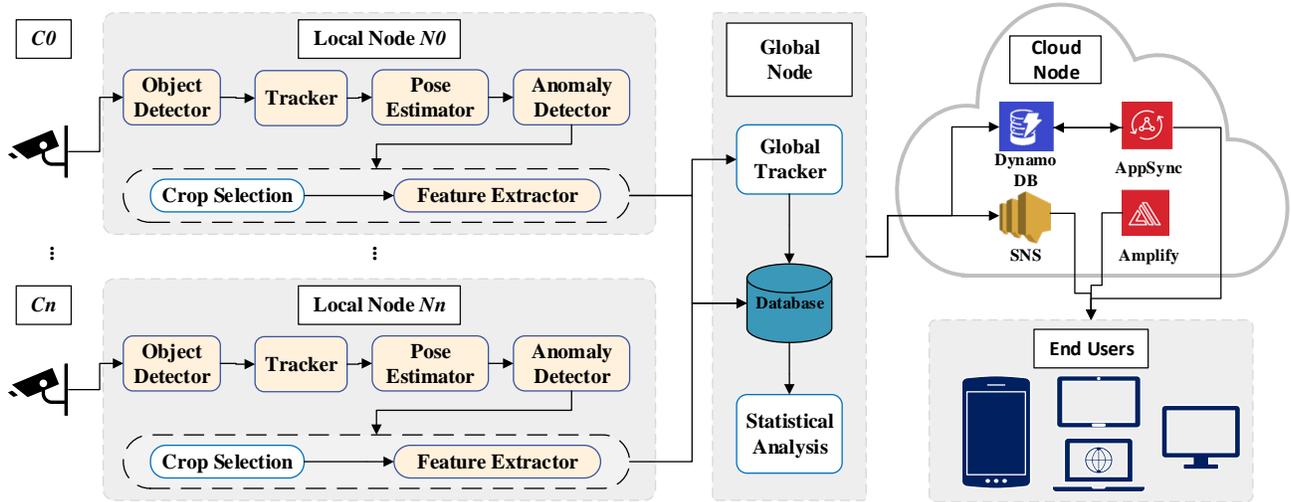}
                    \caption{End-to-end detailed system. \textit{$C_{0}$} represents the camera \textit{ID} and for each camera, one local Node \textit{$N_{0}$} including multi-AI-vision-models pipeline is assigned with. All the local nodes send processed data to one Global Node database, and Global Node will re-identify the human track ID based on the feature extractor data. The statistical analyzer analyzes all the data in the database stored across all the cameras and communicates the results with Cloud Node. Cloud-native services are utilized to host the end users' applications.}
                    
                \label{fig:pipe}
\end{figure*}
Suspicious object detection is crucial for security purposes in real-life applications; however, large object classification datasets like COCO do not include harmful objects. To achieve improved object detection results, a study of Real-Time abnormal object detection \cite{38} shows how to train and implement the abnormal object detection model into smart cities with reaching about 90 percent accuracy\cite{35}\cite{36}. Another paper published in 2023, which discussed person re-identification using deep learning-assisted methods\cite{40} focuses on learning spatial and channel attention between different views of the same object to have a machine learning-related solution for better re-identification scores of 24.6 percent and 54.8 percent.

Despite the advancements made in the aforementioned related works, there remains a lack of substantial testbed support to evaluate the real-world performance of SVS systems. As the motivation for our study, which aims to demonstrate the capabilities of a state-of-the-art SVS system in a public area, such as a community college, we are going to present our achievements in the following sections.

\section{System Features}
In this case study, the system comprises two distinct servers, namely one edge and one cloud. The edge node hosts a sophisticated deep learning pipeline, which employs multiple algorithms, including one high-level action anomaly detection model, to potentially identify suspicious objects and anomalous activities within the area under surveillance. This deep learning pipeline is designed to operate seamlessly with the onsite surveillance cameras, enabling real-time detection of potential security threats. The cloud node receives and processes all the data gathered from the local node for each camera node in a centralized manner. Leveraging cloud computing capabilities, this node performs data analysis and generates alerts to inform end-users of two kinds of potential anomalies: object and behavioral.

 Fig \ref{fig:pipe} shows the AI algorithm pipeline on the local node and global node used for communication with the cloud node in the edge.\cite{2} For each camera one AI algorithm pipeline is assigned for processing. The local node deep learning pipeline commences by extracting images from the surveillance cameras and downsampling their resolution to 720p to ensure consistent processing rates across all cameras in the loop. The images are then batched into sets of 30 frames to match the window size of the high-level tasks later on. 
 
\begin{table*}[]
\centering
\caption{Detailed System hardware configuration for the local server}
\label{tab:hardware-config}
\begin{tabular}{c|cccc|cccc}
\rowcolor{DarkGray}
Name &
  \multicolumn{4}{c|}{Processor} &
  \multicolumn{4}{c}{GPU} \\
 \rowcolor{Gray}&
% \rowcolor{Gray}
  \multicolumn{1}{c|}{Model} &
  \multicolumn{1}{c|}{Cores} &
  \multicolumn{1}{c|}{Clock Speed} &
  Memory &
  \multicolumn{1}{c|}{Model} &
  \multicolumn{1}{c|}{CUDA Cores} &
  \multicolumn{1}{c|}{Boost Clock} &
  VRAM \\ \hline
Local Node &
  \multicolumn{1}{c|}{1 x AMD EPYC} &
  \multicolumn{1}{c|}{16} &
  \multicolumn{1}{c|}{1500 MHz} &
  252 GB &
  \multicolumn{1}{c|}{4 x RTX 3090} &
  \multicolumn{1}{c|}{10496} &
  \multicolumn{1}{c|}{1700 GHZ} &
  24 GB
\end{tabular}
\end{table*}
The batched image data is initially fed into an object detector\cite{30}, which can detect and localize objects of interest, such as people, vehicles, and especially suspicious items (including guns and knives). Object anomaly detection is detected and notified at this stage for our system. Meanwhile, the location data of persons is transmitted to the next stage, which can track and match each detection \cite{31} in previous images. Tracklets are created within the tracker and processed into the format of local IDs assigned to each person's detections. The tracklets data will be used to allocate different people for pose estimation \cite{33} in the next stage. As mentioned earlier, action anomaly detection in the future is a high-level task requiring human pose data. After the persons' local IDs are identified, locational coordinates are used to extract 2D skeletons for the GEPC anomaly detector \cite{32}, which looks at each person's 2D pose-estimation data in a window size of 30 frames with a stride of 20 frames in our edge node system. This detector generates scores individually for each person by looking at the person's movements in batches. Based on the scores for each detection, a scene anomaly score is calculated at the end. For each batch passing through the pipeline, any batch containing person detections is generated with one scene anomaly score for every frame within the batch.
 After the anomaly score is generated, the batch is sent to the next stage, including crop selection, which is a mathematical method to select the best frame with the richest information after comparing it with other frames in the batch to reduce the database load. One feature extractor algorithm is used to extract each human's features in the frame for human track re-identification across different cameras at the global node.
 
 Finally, all the data generated from a single camera through the pipeline is transferred to the global Node. The global tracker will process all the human features across all the cameras in the system to track humans and the database stores all the data processed from the pipelines for historical recording purposes. In order to reduce the load on the cloud node side, all the data sent to the cloud node is run through statistical analysis to generate better statistical anomaly data and action anomaly data. More details can be found in \cite{2}

\subsection{Cloud Node}

In this section, we the cloud node of the utilized system. We are using various cloud-native services that provide robust data storage, management, user management solutions, and API generator services\cite{26}. Figure \ref{fig:pipe} shows the architecture of the cloud node in this setup. The architecture is designed with scalability in mind, and we use different Amazon Web Services (AWS) services to enable the system to meet its objectives.

The system architecture is based on the idea to reduce the latency in sending notifications to the end users as much as possible. A cloud-based service sends push notifications to end-users devices to ensure that users are promptly notified in emergencies. AWS Simple Notification Service (SNS) is leveraged to create necessary topics and associated messages\cite{29}. Specific topics and messages are generated on the AWS SNS based on the type of emergency cases, such as detecting suspicious objects, and behavioral anomalies. In the global node analysis section, emergencies are identified and directed toward their designated topic. The AWS SNS uses the JSON protocol to communicate with the end-user devices and disseminate the notifications accordingly. The notifications could be delivered by the end user based on their subscription method; which could be through an email, text message, and pop-up notification on the installed application. We opted to send the notifications via email to the end-users to ensure the limitations of the delivery method, such as service outages or network outages are not posing latency to the whole system. 

The AWS DynamoDB database stores data on the cloud to enable the system to send real-time data; this database enables application developers to query the stored data with the key-value attribute\cite{27}. We use two types of tables to store data: tables that store the number of tracked objects across each camera and tables that store the analytical results associated with each camera over time. These tables are differentiated through the key-value attribute. We use timestamps and camera ids as the key values, enabling the client to access all data easily. As a result, users can search the database for desired statistics over time and in different locations on the user's device.
An application development service is required to generate the application Programming Interfaces (API) for the smartphone application. We are using the AWS AppSync service. This service creates a GraphQL schema to import data from existing tables. To avoid over-fetching data from the cloud to mobile and improve performance by retrieving data from the cloud efficiently, we use GraphQL. This two-way communication service uses the Hypertext Transfer Protocol Secure (HTTPS) protocol to send data to the end-user's device, enabling end-users to connect to the database and search for the desired statistics\cite{28}.

User management and authentication are other aspects of the mobile application that cloud services handle with AWS Amplify. However, the details of such implementations go beyond the scope of this paper. Overall, the end-to-end system architecture that we have developed provides an efficient and scalable solution for hosting the smartphone application that is used to deliver notifications to the end user.
As discussed earlier, the ultimate objective of this system is securely notifying the end user in case any anomaly is detected. To accomplish this objective, we developed a smartphone application. Through the utilization of this smartphone application, end users are capable of acquiring real-time data, including the number of individuals present at each location, an occupancy indicator based on historical data at each location, a real-time bird's eye view of pedestrians captured by each camera, an occupancy pattern of each location through the generation of a heat map, the type, time, and the number of anomalous behaviors, as well as cumulative data on the total and average number of detected objects associated with each location over time. Moreover, if any anomaly, including behavioral, and suspicious objects, is identified, the end user will receive a notification on their device based on their subscription methods.

\section{Test set up and Results}
\subsection{Test setup}
\begin{figure*}[]
        \centering
               \includegraphics[width=1\linewidth, trim= 0 180 5 120,clip]{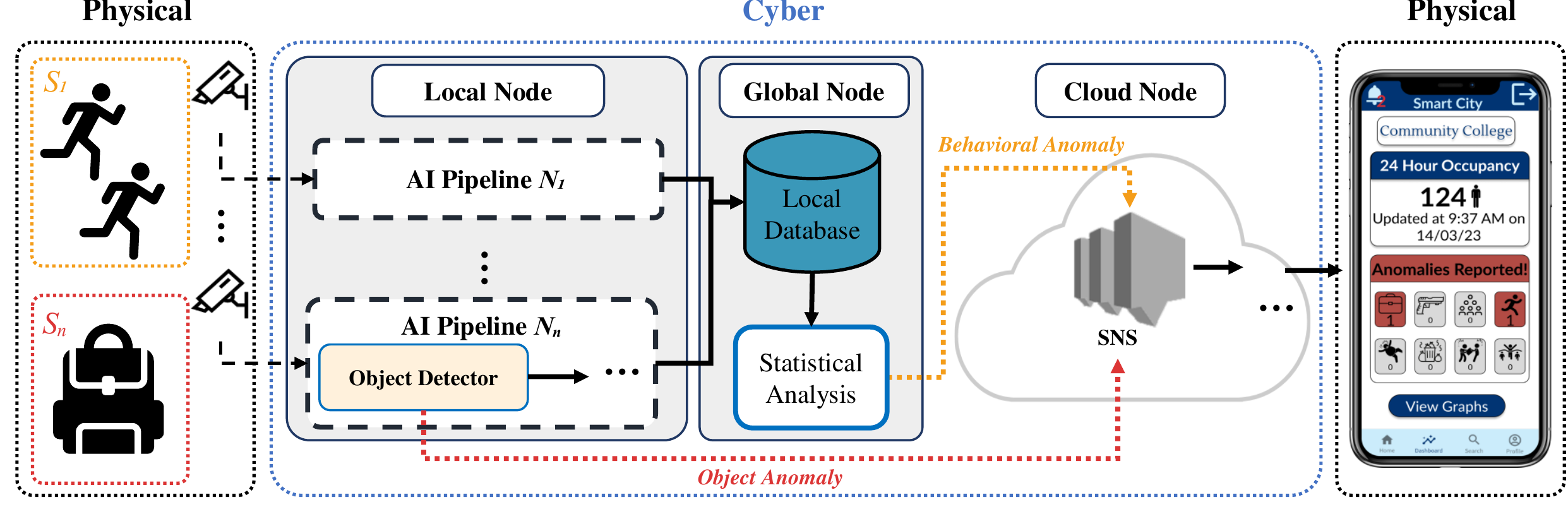}
               \caption{Notification's data flow through Physical-Cyber-Physical set up. \textit{$S_{1}$} represents the first scene that detects a behavior anomaly. The yellow line shows the notification data flow to the SNS service. \textit{$S_{n}$} shows the \textit{$n^{\text{th}}$} scene where a suspicious object has been detected. The red line shows the notification data flow when detecting object anomalies.}

                \label{fig:Notif}
\end{figure*}

We ran the system at a community college as our testbed. In this setup, we use a hardware server configuration consisting of a single AMD EPYC CPU and four RTX 3090 GPUs. More detailed information on the hardware configuration can be found in table \ref{tab:hardware-config}.

For the tests, experiments were conducted under two scenarios to measure the system performance and the latencies in real time.
Under the first scenario, the system's average latency and throughput are measured. The tests were conducted using different numbers of cameras to address the scalability issue of the system. In this scenario, we run the experiments simultaneously with one, two, four, and eight cameras. We also utilized cameras in both indoor and outdoor locations. In the one-camera experiment, we used an indoor camera in a hallway. For the two-cameras experiments, we operated an indoor camera and a parking lot as an outdoor camera. We used two cameras in hallways and two parking lot cameras to run our four-camera experiment. Finally, in the eight-camera experiment setup, we used two parking lot cameras as outdoor cameras. We used cameras in three hallways, two entrances of buildings, and one camera recording vending machine. During the experiments, we kept it continuously running for 20 minutes on campus without controlling the density or behaviors of the pedestrians to ensure that the collected data reflected real-life scenarios. To obtain more latency and throughput results when anomalies are detected, the anomaly object is set as "backpack," which is a more familiar object in a community college. Additionally, the action anomaly threshold is adjusted to the level where all movement detections are treated as anomalies. In this setup, we can simulate how the system performs under computation pressure due to a high-density scene.

The second scenario was developed to measure the Physical-Cyber-Physical (P-C-P) latency, which measures the end-to-end latency from when an anomaly occurs in front of the camera to when the users receive a notification. To ensure the scientific validity of the collected data, the system is configured to report anomaly notifications from only one camera that is simulated in front of it. In contrast, other cameras continue running in parallel without pushing alerts. In this scenario, we run the experiments with one, four, and eight cameras. The choices of the cameras were the same as in the first scenario. Two types of notifications are simulated and reported separately. One type is object anomaly PCP latency, which indicates the time between detecting high-priority object classes and the end user receiving the alert. The other is a behavioral anomaly, which shows the performance of dangerous activities in the scene, such as fighting, chasing, jumping, and falling. We performed three experiments, with one, four, and eight cameras, for each of the two types of anomalies, resulting in a total of 180 data samples to measure the PCP latency of the system in different conditions. 

Figure \ref{fig:Notif} shows a more detailed understanding of our notification data flow with PCP latency. The video will be sent to the AI vision pipeline in the local node as soon as the scene is presented before the surveillance camera. Once suspicious items are detected, the AI pipeline will send a notification through SNS to alert the end users, shown in red dotted line as Object Anomaly path in figure \ref{fig:Notif}. On the other hand, when the camera detects an anomaly scene, the data will pass all pipeline stages and later will be sent to the global node database for more statistical analysis to reduce possible False Negatives (FN) and False Positives (FP) of the system. Once the analyzed anomaly score passes the defined threshold, it will be pushed to the SNS service to send notifications. The paths are represented in figure \ref{fig:Notif} as Behavioral Anomaly with the yellow dotted line.  

\subsection {System Results}

\begin{table}[]
\centering
\caption{Average throughput and latency of local server with different number of cameras}
\label{tab:local_server}
\begin{tabular}{c|ccc|c}
 \rowcolor{DarkGray}
  & \multicolumn{3}{c|}{Latency (s)}                             & FPS   \\
   \rowcolor{Gray}
Nodes & \multicolumn{1}{c|}{Detector stage} & \multicolumn{1}{c|}{Action stage} & Whole system &  \\ \hline
1 & \multicolumn{1}{c|}{1.45} & \multicolumn{1}{c|}{2.51} & 2.86 & 31.37    \\
2 & \multicolumn{1}{c|}{1.46} & \multicolumn{1}{c|}{2.3}  & 2.57 & 31.05 \\
4 & \multicolumn{1}{c|}{1.59} & \multicolumn{1}{c|}{2.38} & 2.7  & 31.11 \\
8 & \multicolumn{1}{c|}{2.71}    & \multicolumn{1}{c|}{4.42}    & 5.32    & 32.41    
\end{tabular}
\end{table}
% \usepackage{multirow}
% \begin{table*}[]
% \centering
% \caption{Average latency of cloud server with different number of cameras}
% \label{tab:cloud_server}
% \begin{tabular}{c|cc|ccc|cc}
%  \rowcolor{DarkGray}
%  &
%   \multicolumn{2}{c|}{DynamoDB Latency (ms)} &
%   \multicolumn{3}{c|}{Notification Latency (ms)} &
%   \multicolumn{2}{c}{AppSync Latency (ms)} \\
%    \rowcolor{Gray}
% Nodes &
%   \multicolumn{1}{c|}{GetItem} &
%   PutItem &
%   Object &
%   Action &
%   Statictical &
%   \multicolumn{1}{c|}{Action} &
%   Statistical \\ \hline
% 1 &
%   \multicolumn{1}{c|}{\multirow{4}{*}{14.6}} &
%   \multirow{4}{*}{17.5} &
%   137 &
%   142 &
%   41 &
%   \multicolumn{1}{c|}{\multirow{4}{*}{105}} &
%   \multirow{4}{*}{14.4} \\
% 2 & \multicolumn{1}{c|}{} &  & 167 & 176 & 47 & \multicolumn{1}{c|}{} &  \\
% 4 & \multicolumn{1}{c|}{} &  & 200 & 172 & 43 & \multicolumn{1}{c|}{} &  \\
% 8 & \multicolumn{1}{c|}{} &  & -   & -   & -  & \multicolumn{1}{c|}{} & 
% \end{tabular}
% \end{table*}

\begin{table}[]
\centering
\caption{Average latency of cloud server with different number of cameras}
\label{tab:cloud_server}
\begin{tabular}{c|cc|cll|cc}
 \rowcolor{DarkGray}
 &
  \multicolumn{2}{c|}{\begin{tabular}[c]{@{}c@{}}DynamoDB \\ Latency (ms)\end{tabular}} &
  \multicolumn{3}{c|}{\begin{tabular}[c]{@{}c@{}}SNS \\ Latency (ms)\end{tabular}} &
  \multicolumn{2}{c}{\begin{tabular}[c]{@{}c@{}}AppSync \\ Latency (ms)\end{tabular}} \\
   \rowcolor{Gray}
Nodes & \multicolumn{1}{c|}{GetItem} & PutItem & \multicolumn{3}{c|}{}    & \multicolumn{1}{c|}{Action} & Statistical \\ \hline
1 &
  \multicolumn{1}{c|}{\multirow{4}{*}{14.6}} &
  \multirow{4}{*}{17.5} &
  \multicolumn{3}{c|}{140} &
  \multicolumn{1}{c|}{\multirow{4}{*}{105}} &
  \multirow{4}{*}{14.4} \\
2     & \multicolumn{1}{c|}{}        &         & \multicolumn{3}{c|}{172} & \multicolumn{1}{c|}{}       &             \\
4     & \multicolumn{1}{c|}{}        &         & \multicolumn{3}{c|}{186} & \multicolumn{1}{c|}{}       &             \\
8     & \multicolumn{1}{c|}{}        &         & \multicolumn{3}{c|}{150.5}   & \multicolumn{1}{c|}{}       &            
\end{tabular}
\end{table}
To evaluate the system performance, we conducted tests using various numbers of nodes and allowed each run to last for twenty minutes, resulting in a total of 1200 batches. We averaged the metrics for the final 1000 batches within each run to calculate the performance, excluding the beginning batches to account for potential warm-up. Unlike running videos, live surveillance cameras have no cool-down effects from terminating the pipeline system. We are not going to exclude the latest batches from the test. The results are separately presented for both the edge and cloud nodes.

Table \ref{tab:local_server} shows how the local node performs with 1/2/4/8 numbers of cameras. The latencies are recorded within three different stages; the detector stage represents how long it takes the local node to allocate the suspicious object; the action stage represents the time to detect anomalous behavior; and the whole system refers to the latency for the data transferred to database on the global node from one pipeline process. More details are shown in figure \ref{fig:pipe}, and all latencies are presented in seconds.

The average throughput is maintained at the same level, around 31 FPS, which is higher than the input frame rate from the surveillance cameras when utilizing 1/2/4 cameras. In the eight-camera experiment, the average throughput is around 32 FPS. This result could be due to running more cameras with idle detections, time-sensitivity crowd density since the eight-camera experiment was conducted during class time, and GPU performance after fitting two AI vision pipelines into one GPU.
The detector stage latency rises as the number of cameras increases, and the result maintains under 1.6 seconds. Interestingly, the action stage latency with only one camera is higher than with 2 and 4 cameras. This observation can be explained by the fact that the system is not running at its maximum speed in the single-camera experiment. This result also affects the average latency of the whole system. By increasing the number of cameras to 8, which results in a heavier load on the CPU, the latency results almost doubled compared with 1/2/4 cameras.

Table \ref{tab:cloud_server} represents how the cloud node latencies will change with the different number of cameras. DynamoDb, SNS, and AppSync services are the cloud-native services that might pose latency to the system. The table \ref{tab:cloud_server}  displays a cloud node's average latency in four different experiments, where each experiment was conducted for 20 minutes using varying numbers of cameras (1, 2, 4, and 8). The latency measurements were extracted using AWS CloudWatch. The results indicate that the average latencies for the four experiments are almost in the same range, which suggests that the SNS latencies are not correlated with the number of cameras. For instance, the average SNS service latency ranged from 140 milliseconds (ms) for one camera to 186 ms in the four-camera setup; while in the eight-camera experiment, we experienced 150.5 ms latency. It is important to note that we observed a maximum of 2 seconds of latency for one data point. The results show that network issues cause SNS latencies. At the same time, the average latency for different numbers of cameras was around 14.6 ms for receiving the data points from the global node denoted as GetItem and 17.5 ms to show the items on the tables called PutItem operation in DynamoDB. The average latency for AppSync was around 105 ms for presenting behavioral anomalies on the application. While the average latency for AppSync to generate the statistical analysis APIs is 14.4 ms. It is due to the fact that the smartphone application uses more complicated user interface elements like color codes to represent behavioral anomalies. 

It was observed that DynamoDB and AppSync were independent of the number of cameras running. They use the data points pushed from the global node, regardless of the number of cameras. Therefore, their latency is not correlated with the number of cameras. 

\subsection {Physical-Cyber-Physical Results}
\begin{table}[]
\centering
\caption{Average P-C-P latency with different number of cameras}
\label{tab:PCP}
\begin{tabular}{c|cc}
 \rowcolor{DarkGray}
     & \multicolumn{2}{c}{Physical-Cyber-Physical Latency (s)} \\
      \rowcolor{Gray}
Node & \multicolumn{1}{c|}{Object Anomaly}   & Behaviour Anomaly  \\ \hline
1    & \multicolumn{1}{c|}{5.3}              & 7.3             \\
4    & \multicolumn{1}{c|}{5.78}              & 7.63            \\
8    & \multicolumn{1}{c|}{11.11}                & 20.78              
\end{tabular}
\end{table}
The average P-C-P latency is shown in table \ref{tab:PCP}. All these records are calculated manually with a timer, which causes some errors in measuring the accurate time. As is shown, the results are influenced by the increment of camera numbers running in parallel. Although running four cameras does not influence the average PCP latency compared to the one-camera experiment dramatically, the average object anomaly PCP latency increased from 5.5 s in the one-camera experiment to 11.11 s when running eight cameras in parallel. The average behavioral anomaly latency suggests the same observation. The average behavioral anomaly PCP for one camera is 7.3 s, which increased to 20.78 s in the eight-camera experiment. These similar observations could be explained by the increased density across all cameras while running the experiment, the whole system's latency increment according to table \ref{tab:local_server}, and the overhead caused by multiple tasks running simultaneously from loading the camera footage to publish the notification message on the user's device.
On the other hand, table  \ref{tab:PCP} indicates that the average object detection PCP latency for one camera is around 5.5 seconds while the behavioral anomaly PCP latency shows a 2 seconds increment. This observation is also similar among different experiments. The reason is that the action anomaly notification needs to run through more stages and run through and recorded in the global node before pushing to the SNS service at the cloud node, as shown in figure \ref{fig:Notif}.  

\begin{figure}[]
        \centering
               \includegraphics[width=1\linewidth, trim= 0 0 0 0,clip]{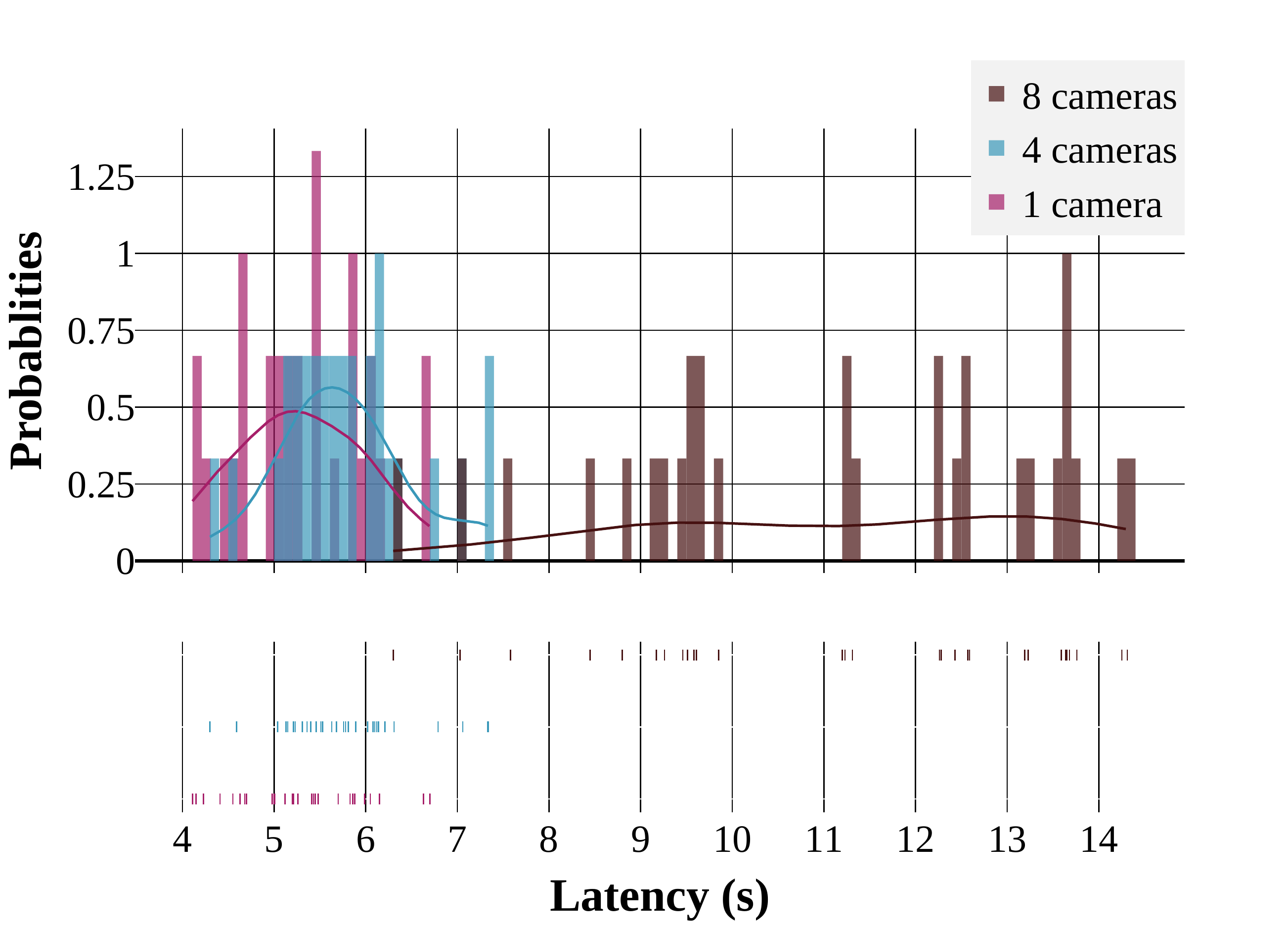}
                    \caption{Object Anomaly Latency Rug plot with each runs during PCP test. 30 data points for each camera number }
                    
                \label{fig:OA_graph}
\end{figure} 

\begin{figure}[]
        \centering
               \includegraphics[width=1\linewidth, trim= 0 0 0 0,clip]{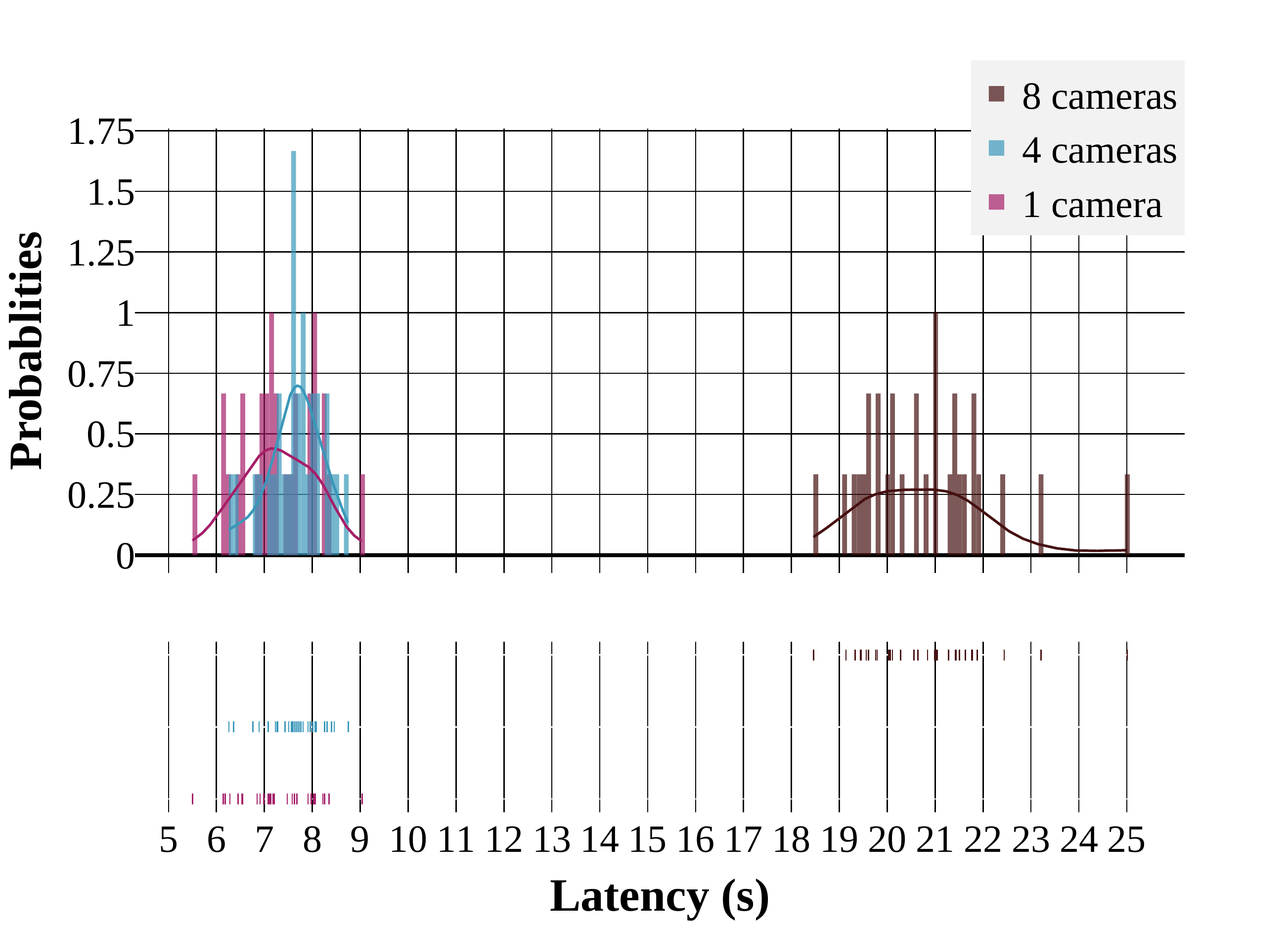}
                    \caption{Behavior Anomaly Latency Rug plot with each runs during PCP test. 30 data points for each camera number }
                    
                \label{fig:BA_graph}
\end{figure}

Figures \ref{fig:OA_graph} and \ref{fig:BA_graph} represent the distributions of the 180 collected data points in our different PCP experiments. Figure \ref{fig:OA_graph} shows that while the data points in the object detection PCP at the single-camera and four-cameras have a lower range of data, the eight-camera experiment's data points are more scattered. Table  \ref{tab:PCP detail} reports the standard deviation of the object detection PCP experiments as 0.651, 0.85, and 2.06 for one, four, and eight cameras, respectively. This observation also can be explained by the reasons mentioned in the former paragraph. We can also see in table \ref{tab:PCP detail} that the minimum latency observed in detecting a suspicious object is 4.11s and 4.3s in the one-camera and the four-camera, respectively.

In contrast, we could not observe any latency lower than 6.3 s when running eight cameras. On the other hand, it is shown in figure \ref{fig:BA_graph} that the eight-camera experiment latency data points in detecting anomalous behaviors are less scattered compared to the object detection PCP experiment with a standard deviation of 1.54. In these experiments, we observed a maximum latency of 25.01 s when running eight cameras. Table \ref{tab:PCP detail} shows the detailed statistical information of these experiments.  
\begin{table}[]
\centering
\caption{Statistical PCP latency data for object and behaviour anomaly for different number of cameras}
\label{tab:PCP detail}
% \small 
\setlength\tabcolsep{5pt}
% \resizebox{\textwidth}{!}
\begin{tabular}{c|ccc||ccc}
 \rowcolor{DarkGray}
  & \multicolumn{3}{c||}{Object Anomaly}                           & \multicolumn{3}{c}{Behaviour Anomaly}                          \\
   \rowcolor{Gray}
Nodes &
  \multicolumn{1}{c|}{Min (s)} &
  \multicolumn{1}{c|}{Max (s)} &
  \begin{tabular}[c]{@{}c@{}}Standard\\ deviation\end{tabular} &
  \multicolumn{1}{c|}{Min (s)} &
  \multicolumn{1}{c|}{Max (s)} &
  \begin{tabular}[c]{@{}c@{}}Standard\\ deviation\end{tabular} \\ \hline
1 & \multicolumn{1}{c|}{4.11} & \multicolumn{1}{c|}{6.7}   & 0.65 & \multicolumn{1}{c|}{5.5}   & \multicolumn{1}{c|}{9.04}  & 0.77 \\
4 & \multicolumn{1}{c|}{4.3}  & \multicolumn{1}{c|}{7.34}  & 0.85 & \multicolumn{1}{c|}{6.26}  & \multicolumn{1}{c|}{8.75}  & 0.61 \\
8 & \multicolumn{1}{c|}{6.3}  & \multicolumn{1}{c|}{14.31} & 2.06 & \multicolumn{1}{c|}{18.46} & \multicolumn{1}{c|}{25.01} & 1.54
\end{tabular}
\end{table}

\section{Conclusion and Discussion}
SVS systems have become an efficient tool for monitoring public safety and security, with the development of machine learning algorithms improving their reliability. While the market for this technology is expected to grow significantly in the coming years, scalability, latency, throughputs, and privacy concerns must be addressed to deploy SVS systems in practical settings. Involving the community in planning and implementing SVS systems can ensure ethical use and protect individuals' privacy. The study in this paper evaluated the latency and throughput performance of an SVS system in detecting anomalies in a campus setting, demonstrating its ability to provide timely alerts to end-users. Our results show that while the local node's latency slightly increased from 2.86 seconds to 5.32 seconds, running one camera to eight cameras, the throughput performance does not change much in one node setup compared to eight nodes, which shows that the system is performing promising in scale. Moreover, the physical-cyber-physical experiments show  the average latency in detecting suspicious objects and behavioral anomalies while running one camera are 5.3 and 7.3 seconds, respectively. Running eight cameras resulted in 11.11 and 20.78 seconds of average latency in detecting object and behavioral anomalies, respectively.   

While our study shows that AI-driven SVS systems can be implemented in the real world to improve surveillance systems while addressing privacy concerns, there are future research directions that can improve these systems' performance. Besides algorithmic and model improvements, engaging the local community and involving them in the planning and implementation of SVS systems is essential to establish trust and acceptance of such systems. Future research should address scalability and privacy concerns and explore the "community in the loop" concept's effectiveness. Evaluating SVS system performance in different environments and scenarios can also address other challenges specific to the domain shift, improving their reliability and efficiency. Overall, this study highlights the potential of SVS systems in enhancing public safety and security and provides practical insights in designing and implementing such systems.

\section*{Acknowledgment}
This research is supported by the National Science Foundation (NSF) under Award No. 1831795.

\end{document}